\documentclass[11pt,a4paper]{article}
\pdfoutput=1
\usepackage{jinstpub}
\usepackage{subcaption}
\usepackage{blkarray, bigstrut}

\graphicspath{{./graphics/}}

\DeclareMathOperator*{\E}{\mathbb{E}}

\title{Machine Learning on data with sPlot background subtraction}
\author[a]{M.~Borisyak}
\author[a,b,1]{N.~Kazeev%
\note{Corresponding author.}}
\affiliation[a]{Laboratory of Methods for Big Data Analysis, National Research University Higher School of Economics, 11, Pokrovsky boulevard, Moscow 109028, Russia}
\affiliation[b]{Sapienza University of Rome, Piazzale Aldo Moro, 5, 00185 Roma RM, Italy}
\emailAdd{nikita.kazeev@cern.ch}
\abstract{Data analysis in high energy physics often deals with data samples consisting of a mixture of signal and background events. The sPlot technique is a common method to subtract the contribution of the background by assigning weights to events. Part of the weights are by design negative. Negative weights lead to the divergence of some machine learning algorithms training due to absence of the lower bound in the loss function. In this paper we propose a mathematically rigorous way to train machine learning algorithms on data samples with background described by sPlot to obtain signal probabilities conditioned on observables, without encountering negative event weight at all. This allows usage of any out-of-the-box machine learning methods on such data.}
\arxivnumber{1905.11719}
\begin{document}
\maketitle

\section{Introduction}
Experimental data obtained in high energy physics experiments usually consists of contributions from different event sources. We consider the case, where the distribution of some variables is known for each source and call these variables discriminative. Usually, the variable is the reconstructed invariant mass, and the probability densities are estimated by a maximum likelihood fit. The distribution of other (control) variables also presents interest, as an analysis quality check or a training sample of a machine learning algorithm. The sPlot technique \cite{splot} allows to reconstruct the distribution of the control variables, provided they are independent of the discriminative variables. sPlot assigns weights (sWeights) to events, some of them negative. This does not present a problem for simple one-dimensional analysis tools, like histograms, but is an obstacle for multivariate machine learning methods that require the loss function to be bounded from below.

This paper is structured as following. In section \ref{sPlot} we briefly introduce the sPlot technique. In section \ref{previous} we describe the literature concerning machine learning on data weighted with the sPlot. In section \ref{sec:negative-weights} we discuss the implications of negative event weights on training of machine learning algorithms. In section \ref{we} we propose methods to robustly obtain class probabilities conditioned on the control variables. In section \ref{sec:experiments} we present experimental results that demonstrate practical viability of the proposed method.

\section{sPlot}
\label{sPlot}
\begin{figure}[h]
    \begin{subfigure}{.5\textwidth}
        \centering
        \includegraphics[width=\textwidth]{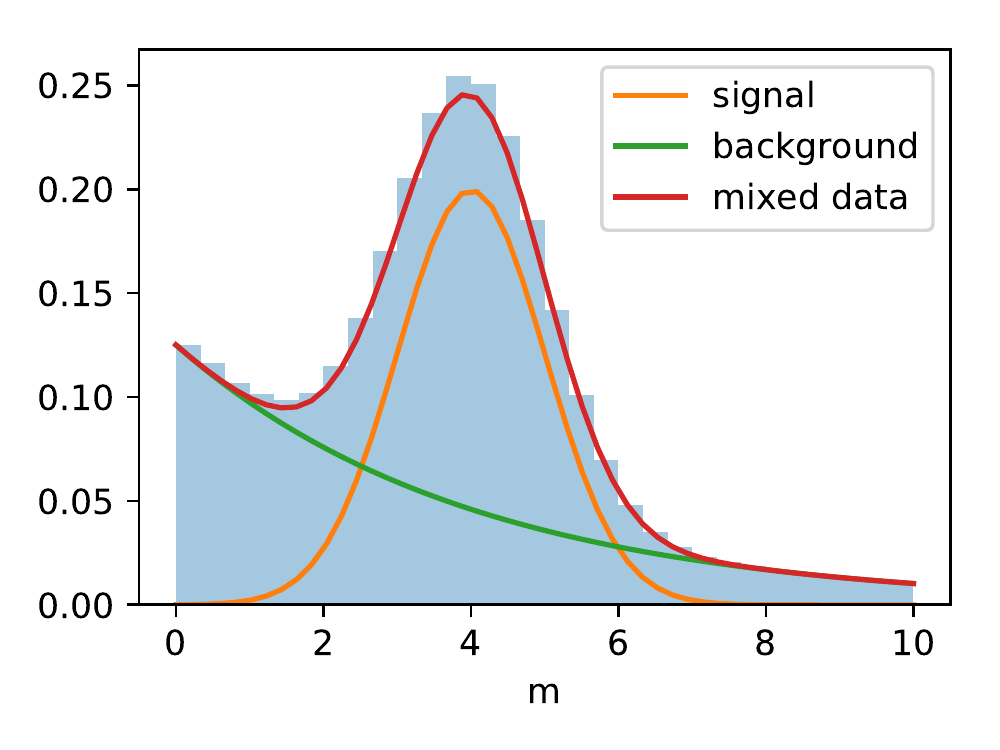}
    \end{subfigure}
    \begin{subfigure}{.5\textwidth}
        \centering
        \includegraphics[width=\textwidth]{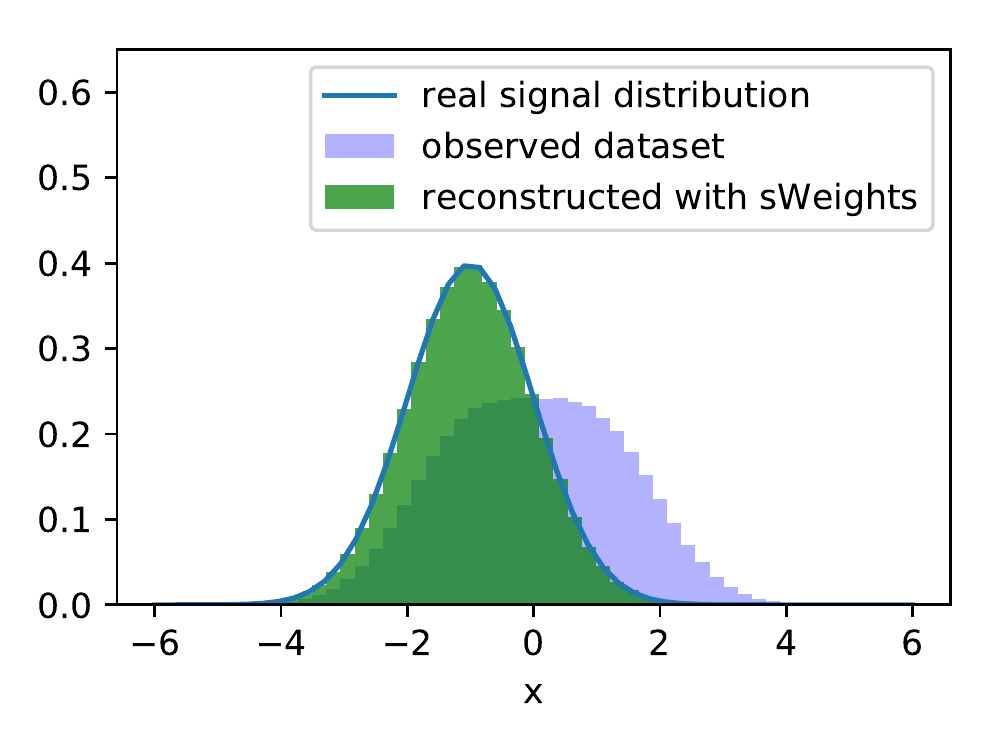}
    \end{subfigure}
    \caption{An example of sPlot application. To the left: the known distributions of $m$. To the right: the mixture and reconstructed distribution of $x$.} \label{fig:sWeights}
\end{figure}

The full derivation of sPlot is present in \cite{splot}. Here we briefly recite the key formulas. Take a dataset populated by events from $N_s$ sources. Let $p_k(m)$ be probability density function of the discriminative variable $m$ of the $i$-th species; $N_k$ the number of events expected on the average for the $k$-th species; $N$ -- the total number of events; $m_e$ -- the value of $m$ for the $e$-th event. Define matrix $\mathbf{V}$:
\begin{equation}
  \mathbf{V}_{nj}^{-1} = \sum_{e=1}^N\frac{p_n(m_e)p_j(m_e)}{\left(\sum_{k=1}^{N^s}N_k p_k(m_e)\right)^2}
\end{equation}
The sWeight for the $e$-th event corresponding to the $n$-the species is obtained using the following transformation:
\begin{equation}
\label{eq:sweight}
  \text{sWeight}_n(m_e) = \frac{\sum_{j=1}^{N_s}\mathbf{V}_{nj}p_j(m_e)}{\sum_{k=1}^{N_s}N_k p_k(m_e)}
\end{equation}
If the dataset is weighted with sWeights, the distribution of the control variables will be an unbiased estimate of the corresponding pure species. In the rest of the paper we deal with the two species scenario, named signal and background. An example of sPlot application is present on figure \ref{fig:sWeights}.

\section{Related work} \label{previous}
There are several works concerning machine learning and sWeights: \cite{Keck:2017mui, lipp2015splot, martschei2012advanced, hoecker2007tmva}. They propose a training procedure for the case where a classifier is desired to separate the same signal and background that are defined by the sPlot. Take each event twice, once as signal, once as background with the corresponding sWeights, then train the classifier as usual. The works also demonstrate practical viability of using machine learning methods on data weighted with sPlot. They, however, do not attempt to analyze the core issue of negative weights impact on machine learning algorithms, liming themselves to requiring that the classifier supports negative weights.

Another possible approach is suggested by \cite{metodiev2017classification} -- classification without labels (CWoLa). It is possible to use the discriminative variable to define regions with different signal/background proportions, then train a classifier to separate them, without using any weights or true labels. The authors prove that the optimal classifier for this case would also be the optimal classifier for signal against background. This method ignores the per-event probability information. In our experiments (presented in section \ref{sec:experiments}), CWoLa has worse quality on a real finite dataset than the methods that use this information.

\section{The problem of negative weights} \label{sec:negative-weights}
\begin{figure}[h]
    \centering
    \includegraphics[width=\textwidth]{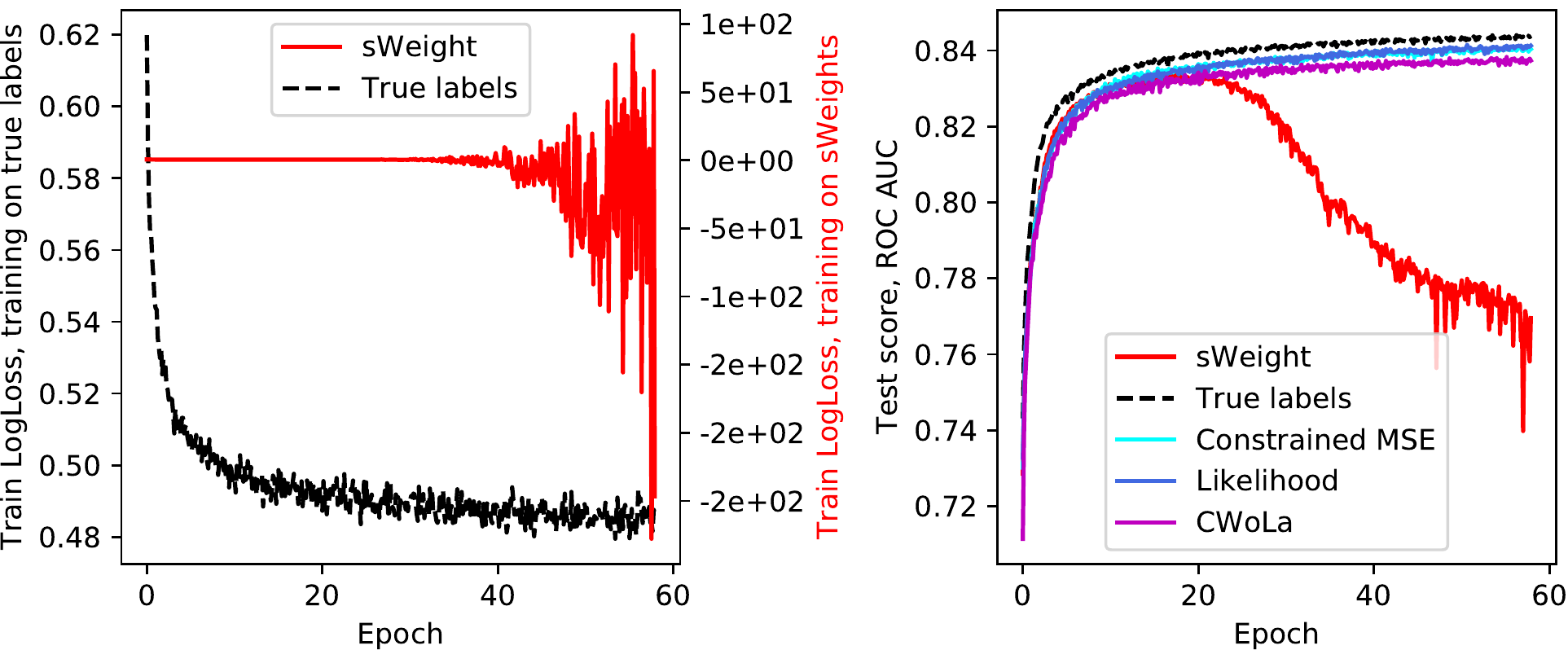}
    \caption{Learning curves of a neural network trained on the Higgs dataset using the true labels and the artificially introduced sWeights. Likelihood and Constrained MSE methods are described in section \ref{we}, CWoLa in \cite{metodiev2017classification}, the dataset in section \ref{sec:experiments}, and the network in appendix \ref{sec:model_parameters}.}
    \label{fig:learning-curve}
\end{figure}

Let us consider an event with positive signal weight $w_s > 0$ and negative background weight $w_b < 0$ and the classic cross-entropy loss function:
\begin{equation}
    L = -w_s\log(p_s) - w_b\log(1 - p_s),
\end{equation}
where $p_s$ is the model output -- the predicted probability of this event being signal.
\begin{equation}
    \lim_{p_s\rightarrow 1} L = -\left( - |w_b|\right) \lim_{p_s\rightarrow 1} \log(1 - p_s) = -\infty.
\end{equation}
Thus, directly incorporating sWeights into the cross-entropy loss causes it to lose the lower bound. The same holds for the mean squared error and other losses without an upper bound in the unweighted case. Training most machine learning algorithms is an optimization problem, and, for some algorithms, such as a large-capacity fully-connected neural network, negative event weights make this optimization problem ill-defined, as the underlying optimization target loses the lower bound as well. An example illustrating diverging training is present on figure \ref{fig:learning-curve}. The model training using the sWeights as event weights quickly diverges in contrast to the same model training using the true labels or our losses. The model trained on true labels does not even start to overfit -- the test score keeps climbing, while the test score of the model trained with sWeights as event weights drops dramatically. As expected, the test ROC AUC for the model that was trained using the true labels is the best among all methods.

However, most machine learning algorithms do not blindly minimize the loss value on the training dataset, as this is likely to lead to overfitting. They add various regularization terms to the optimized functional that, in general, penalize model complexity or overconfidence. We are not aware of peer-reviewed literature exploring the impact of regularization on learning with negative weights. One such technique appears in discussions within the High Energy Physics community \cite{github-negweights}. They propose avoiding the unbounded loss by requiring leafs of decision tree to have positive total weight. It is also possible to regularize neural networks into having bounded loss. For example, by using the L2 regularization on weights and taking the root of the degree equal to the number of layers plus one from the cross-entropy loss. 

Nevertheless, it is unclear how negative weights combined with such regularization would affect classifier performance. A more detailed study falls outside of the scope of this paper, which presents a principled approach to avoid the problem entirely.

\section{Our approaches} \label{we}
\subsection{sWeights averaging (Constrained MSE)} \label{sec:ConstrainedMSE}

Let $m$ be the variable that was used to compute the sWeigths, $x$ the rest of the variables. It can be shown that:
\begin{equation}
    \E_{m}\left[ w(m) \mid x \right] = \frac{p_\text{signal}(x)}{p_\text{mix}(x)}.
    \label{eq:average-sWeight}
\end{equation}
The formal proof is available in appendix \ref{sec:proof-MSE}. Notice, that the right-hand side of (\ref{eq:average-sWeight}) is the optimal output of a classifier, while the left-hand side can be estimated by a regression model. Our first proposed approach is to perform mean-square regression directly on sWeights. Since the optimal output lies in $[0, 1]$, one can easily avoid a priori incorrect solutions by, for example, applying sigmoid function to the model output. The resulting loss function is the following:


\begin{equation}
    L = \sum_i \left(w_i - \frac{e^{f_\theta(x_i)}}{1+e^{f_\theta(x_i)}}\right)^2, \label{eq:constrainedMSE}
\end{equation}
where $w_i$ is the sWeight and ${f_\theta(x_i)}$ is the model output. This loss has been implemented for the CatBoost machine learning library \cite{prokhorenkova2018catboost} and is available on GitHub.\footnote{\url{\detokenize{https://github.com/kazeevn/catboost/tree/constrained_regression}}}

\subsection{Exact maximum likelihood} \label{sec:likelihood}
Alternatively, one can invoke Maximum Likelihood principle and avoid the sPlot technique altogether. This leads to the following loss function (derivation is in \ref{sec:proof-Likelihood}):
\begin{equation}
    L(\theta) = -\sum_i \log \left[ f_\theta(x_i) p_\text{signal}(m_i) + (1 - f_\theta(x_i)) p_\text{background}(m_i) \right],
    \label{eq:exactloss}
\end{equation}
where $f_\theta(x_i)$ is the output of the model; $p_\text{signal}(m_i)$ and $p_\text{background}(m_i)$ are the probability densities of the signal and background $m$ distributions.

Note, that by substituting $p_\text{signal|background}(m_i)$ by the class indicator ($y_i = 1$ if $x_i$ is a signal sample, $y_i = 0$ otherwise) in loss \eqref{eq:exactloss}, a conventional expression for cross-entropy loss can be obtained.

\section{Experimental evaluation} \label{sec:experiments}
To demonstrate viability of our methods on practical problems, we tested them on the UCI Higgs dataset  \cite{baldi2014searching, uci}. We used neural network and gradient boosting models, their detailed description is in \ref{sec:model_parameters}. The dataset is the largest open dataset from the field of High-Energy Physics. It has 28 tabular features. We split it into train and test parts containing $8.8\cdot 10^6$ and $2.2\cdot 10^6$ events respectively. The dataset is labeled and it does not feature sWeights, so we introduced them artificially. For both signal and background events we added a virtual "mass" distributed as shown on figure \ref{fig:sWeights} and used it to compute sWeights. We also compared with the CWoLa method \cite{metodiev2017classification} by splitting the data into a signal-majority and a background-majority regions. The signal region was chosen with center at $m=4$ (center of the signal $m$ distribution) and width so that it would include half of the events.

\begin{figure}
    \begin{subfigure}[b]{0.5\textwidth}
        \includegraphics[width=\textwidth]{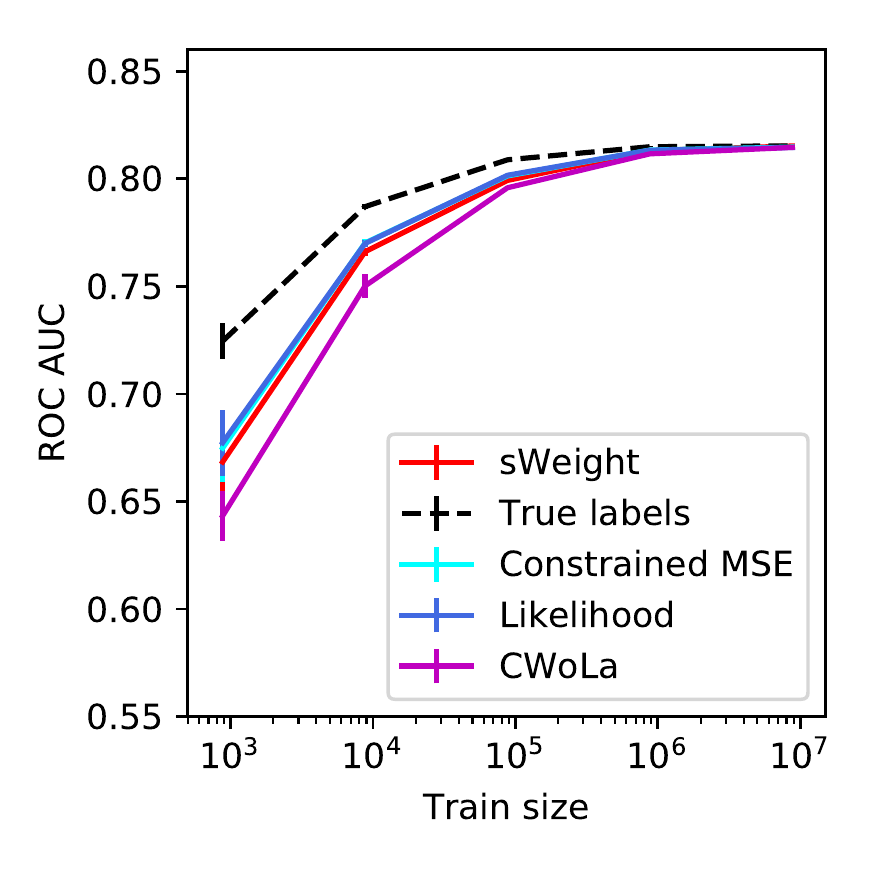}
        \caption{CatBoost}
    \end{subfigure}
    \begin{subfigure}[b]{0.5\textwidth}
        \includegraphics[width=\textwidth]{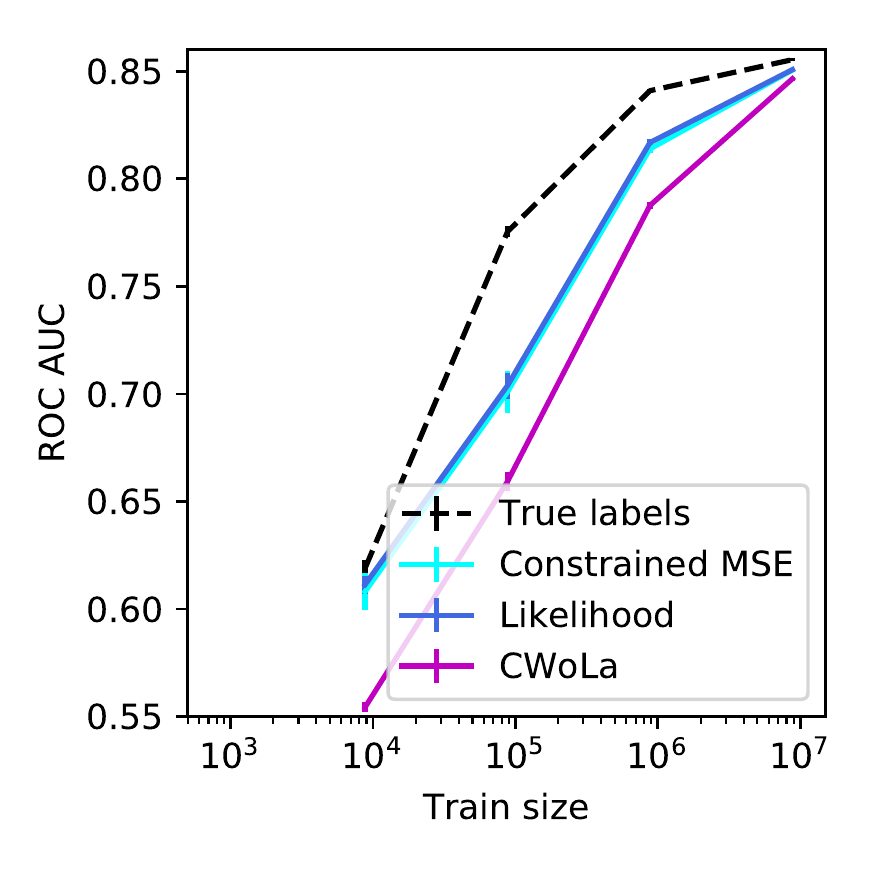}
        \caption{Neural network}
    \end{subfigure}
    \caption{Experimental evaluation of performance of different loss functions on the Higgs dataset as a function of train dataset size. sWeight -- cross-entropy weighted with sWeights, it is not reported for the neural network due to divergence of optimization and, hence, highly stochastic nature of the results; True labels -- logloss using the true labels;  Constrained MSE -- our loss function defined by (\ref{eq:constrainedMSE}); Likelihood -- our loss function defined by (\ref{eq:exactloss}); CWoLa -- method from \cite{metodiev2017classification}; }
    
    \label{fig:higgs}
\end{figure}

The results are present on figures \ref{fig:learning-curve} and \ref{fig:higgs}. As expected, training on true labels gives the best performance. Constrained MSE and Likelihood loss functions show the same performance, both methods outperform CWoLa. For Catboost, directly using sWeights as event weights results in stable training and the same performance as in our methods. For neural networks, using sWeights as event weights leads to divergent training. The performance difference between the methods gets smaller with train dataset size increase. The code of the experiments is available on GitHub.\footnote{\url{https://github.com/yandexdataschool/ML-sWeights-experiments}}


\section{Conclusion}
Training a machine learning algorithm on a dataset with negative weights means dealing with a loss that potentially has no lower bound. The implications depend on the algorithm in question. In our experiments, neural network training diverges, while gradient boosting over oblivious decision trees does not.

Our contribution is the two loss functions that allow a machine learning algorithm to obtain class probabilities from background-subtracted data without encountering negative event weight at all. The probability of an event with given control variables values to be signal is the expected sWeight, that can be estimated by a regression with the corresponding loss function (\ref{sec:ConstrainedMSE}). We invoke the Maximum Likelihood principle to construct a loss function that avoids sPlot and associated problems with negative event weights (\ref{sec:likelihood}).


Our work paves a rigorous way to use any machine learning methods on data with sPlot-based background subtraction.

\acknowledgments
Funding: This work was supported by the Russian Science Foundation [grant number 17-72-20127].

Artem Maevskiy for suggestions on improving learning stability; Denis Derkach for suggestions on experiments and text; Stanislav Kirillov for consultation on the CatBoost code; Mikhail Belous for suggestions on text; anonymous JINST reviewer for bringing to our attention the CWoLa paper~\cite{metodiev2017classification}.

\bibliographystyle{JHEP}
\bibliography{bibliography}
\addcontentsline{toc}{section}{References}

\appendix
\section{Proofs}
\subsection{Constrained MSE} \label{sec:proof-MSE}
Let $m$ be the variable that was used to compute the sWeigths, $x$ be the rest of variables and $f(x)$ is any smooth function of $x$.
$$E_{x\sim p_\text{sig}}(f(x)) = \int dx f(x) p_\text{sig}(x)$$

define

$$W(x) = \frac{p_\text{sig}(x)}{p_\text{mix}(x)}$$

Then

\begin{equation}
    E_{x\sim p_\text{sig}}(f(x)) = \int dx f(x) W(x) p_\text{mix}(x) \label{eq:weight-appenix}
\end{equation}

By definition of sWeights:
$$E_{x\sim p_\text{sig}}(f(x)) = \int dx dm \cdot w(m) f(x) p_\text{mix}(x, m),$$

where $w(m)$ is the sWeight.

$$E_{x\sim p_\text{sig}}(f(x)) = \int dx dm \cdot w(m) f(x) p_\text{mix}(x) p_\text{mix}(m|x)$$
Rearrange the multipliers in the double integral:
$$E_{x\sim p_\text{sig}}(f(x)) = \int dx f(x) p_\text{mix}(x) \int dm w(m) p_\text{mix}(m|x)$$

From equation \ref{eq:weight-appenix}
$$\int dx f(x) W(x) p_\text{mix}(x) =  \int dx f(x) p_\text{mix}(x) \int dm w(m) p_\text{mix}(m|x)$$

Therefore,
$$W(x) = \int dm w(m) p_\text{mix}(m|x) = \mathbb{E}_m(\text{sWeight}(x, m))$$

\subsection{Exact maximum likelihood} \label{sec:proof-Likelihood}
Let us denote signal and background classes as $\mathrm{S}$ and $\mathrm{B}$, model parameters as $\theta$. By definition of the log-likelihood $l(\theta)$:
\begin{equation}
    l(\theta) = \sum_i \log p(x_i, m_i \mid \theta) =
        \sum_i \log \left[ p(x_i, m_i \mid \theta, \mathrm{S}) P(\mathrm{S}) + p(x_i, m_i \mid \theta, \mathrm{B}) P(\mathrm{B}) \right]. \label{eq:likelihood}
\end{equation}

Since $x_i$ and $m_i$ are assumed to be independent within individual classes, expression \eqref{eq:likelihood} can be simplified further:
\begin{multline}
    l(\theta) = \\
    \sum_i \log \left[ p(x_i \mid \theta, \mathrm{S}) p(m_i \mid \mathrm{S}) P(\mathrm{S}) + p(x_i \mid \theta, \mathrm{B}) p(m_i \mid \mathrm{B}) P(\mathrm{B}) \right] = \\
    \sum_i \log \left[ P(\mathrm{S} \mid x_i, \theta) p(m_i \mid \mathrm{S}) p(x_i) + p(\mathrm{B} \mid x_i, \theta) p(m_i \mid \mathrm{B}) p(x_i) \right] = \\
    \sum_i \log \left[ P(\mathrm{S} \mid x_i, \theta) p(m_i \mid \mathrm{S}) + P(\mathrm{B} \mid x_i, \theta) p(m_i \mid \mathrm{B}) \right] + \mathrm{const},
\end{multline}

which leads to the following loss function:
\begin{equation}
    L(\theta) = -\sum_i \log \left[ f_\theta(x_i) p(m_i \mid \mathrm{S}) + (1 - f_\theta(x_i)) p(m_i \mid \mathrm{B}) \right]
\end{equation}

\section{Models parameters} \label{sec:model_parameters}
\subsection{Neural Networks}
For the experiments with fully-connected neural networks we use networks with 3 hidden layers: 128, 64, 32 neurons for the experiments with $8.8 \cdot 10^6$ and $8.8 \cdot 10^5$ samples in the training sets, and 64, 32, 16 neurons in cases of $8.8 \cdot 10^4$ and $8.8 \cdot 10^3$ training samples. Model capacity varies to adjust for the low sample sizes.

All networks use leaky ReLu (0.05) activation function. Networks are optimized by adam \cite{kingma2014adam} algorithm with learning rate $2 \cdot 10^{-4}$, $\beta_1=0.9$, $\beta_2=0.999$ for $2.2 \cdot 10^{6}$ steps (32 full 'passes' over the original training sample) with batch size $128$; learning rate is set to the value lower than commonly used ones due to high variance of the gradients for CWoLa.

Each experiment is repeated 5 times with varying training dataset (if subsampled) and initial weights, however, within each experiment networks for different methods have identical conditions: they share initial weights, are trained on the same subsample and use identical sequence of batches.

Due to use of $\log$ function, loss function \eqref{eq:exactloss} might become computationally unstable for networks with large weights. For the experiments with $8.8 \cdot 10^4$ training samples, $l_2$ regularization is introduced to the exact maximum likelihood loss function. This regularization does not affect results in any significant way, besides limiting network weights to computationally stable values.

\subsection{CatBoost}
We use CatBoost with the following parameters: 1000 trees, leaf\_estimation\_method="Gradient", version 0.10.2 with our losses added and check for negative weights removed: \url{\detokenize{https://github.com/kazeevn/catboost/tree/constrained_regression}}. Each experiment is repeated 10 times with varying training dataset (if subsampled).

\end{document}